\documentclass[letterpaper, 10 pt, conference]{ieeeconf}  

\IEEEoverridecommandlockouts                              

\usepackage{caption}
\usepackage{graphicx}
\usepackage{stfloats}
\usepackage{float} 
\usepackage{subcaption}
\usepackage{color}
\usepackage{amsmath}
\usepackage{cite}
\usepackage{makecell}
\usepackage{booktabs}
\usepackage{multirow}
\usepackage{diagbox}
\makeatletter
\let\NAT@parse\undefined
\makeatother
\usepackage{hyperref}
\hypersetup{pdfstartview=FitH,
            colorlinks=true,
            linkcolor=red,
            anchorcolor=blue,
            citecolor=green
            }

\usepackage{colortbl}
\usepackage{amsfonts}
\usepackage[table]{xcolor}
\newcommand{\ignore}[1]{}
\newcommand{\my}{}
\newcommand{\myAbl}{\cellcolor{gray!20}}
\definecolor{cGreen}{RGB}{100,180,100}
\definecolor{cRed}{RGB}{220,50,0}
\definecolor{Klein_Blue}{rgb}{0.0, 0.129, 0.6}
\newcommand{\sotaBest}[1]{\textbf{\textcolor{cRed}{#1}}}
\newcommand{\sotaSecond}[1]{\textcolor{blue}{#1}}
\newcommand{\sotaThird}[1]{\textcolor{cGreen}{#1}}
\newcommand{\nGOT}[1]{\textcolor{gray!60}{#1}}
\newcommand{\bkdot}{\includegraphics[height=0.5em]{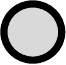}}
\newcommand{\bldot}{\includegraphics[height=0.5em]{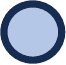}}
\newcommand{\yldot}{\includegraphics[height=0.5em]{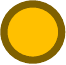}}

\begin{document}
\title{\LARGE \bf
LiteTrack: Layer Pruning with Asynchronous Feature Extraction\\ for Lightweight and Efficient Visual Tracking}
\author{Qingmao Wei$^{1,}$, Bi Zeng$^{1}$, Jianqi Liu$^{1}$*,  Li He$^{2}$, Guotian Zeng$^{1}$
\thanks{*Corresponding author}
\thanks{$^{1}$Q. Wei, B. Zeng, J. Liu, and G. Zeng are with the School of Computer, Guangdong University of Technology, Guangzhou 510006, China. {\tt\small liujianqi@ieee.org}}%
\thanks{$^{2}$L. He is with the Department of Electronic and Electrical Engineering, Southern University of Science and Technology, Shenzhen 518055, China.}%
}


\maketitle
\thispagestyle{empty}
\pagestyle{empty}

\begin{abstract}
	The recent advancements in transformer-based visual trackers have led to significant progress, attributed to their strong modeling capabilities. However, as performance improves, running latency correspondingly increases, presenting a challenge for real-time robotics applications, especially on edge devices with computational constraints. In response to this, we introduce LiteTrack, an efficient transformer-based tracking model optimized for high-speed operations across various devices.  It achieves a more favorable trade-off between accuracy and efficiency than the other lightweight trackers. The main innovations of LiteTrack encompass: 1) asynchronous feature extraction and interaction between the template and search region for better feature fushion and cutting redundant computation, and 2) pruning encoder layers from a heavy tracker to refine the balnace between performance and speed. As an example, our fastest variant, LiteTrack-B4, achieves 65.2\% AO on the GOT-10k benchmark, surpassing all preceding efficient trackers, while running over 100 \emph{fps} with ONNX on the Jetson Orin NX edge device. Moreover, our LiteTrack-B9 reaches competitive 72.2\% AO on GOT-10k and 82.4\% AUC on TrackingNet, and operates at 171 \emph{fps} on an NVIDIA 2080Ti GPU. The code and demo materials will be available at \href{https://github.com/TsingWei/LiteTrack}{https://github.com/TsingWei/LiteTrack}.

\end{abstract}

\section{Introduction}

Visual object tracking is a fundamental task in computer vision, which aims to track an arbitrary object given its initial state in a video sequence. In recent years, with the development of deep neural networks~\cite{AlexNet,ResNet,googlenet,2017Attention}, tracking has made significant progress. In particular, the utilization of transformers~\cite{2017Attention} has played a pivotal role in the development of several high-performance trackers~\cite{TransT,Stark,wang2021transformer,yu2021high,mixformer,ostrack,SeqTrack}. Unfortunately,  a majority of recent research efforts~\cite{SiameseRPN,DiMP,TransT} has concentrated solely on achieving high performance without considering tracking speed.

While these state-of-the-art trackers might deliver real-time performance on powerful GPUs, their efficiency diminishes on devices with limited computational resources. For instance, ARTrack~\cite{ARTrack}, considered as a top-tier tracker, reaches a tracking speed of 37 frames per second (\emph{fps}) on the NVIDIA RTX 2080Ti GPU but drops to 5 \emph{fps} on the Nvidia Jetson Orin NX, a common edge device. This underscores the pressing need for trackers that effectively strike a balance between performance and speed.


\begin{figure}[tbp]
 \centering
 
 \includegraphics[width=0.98\linewidth]{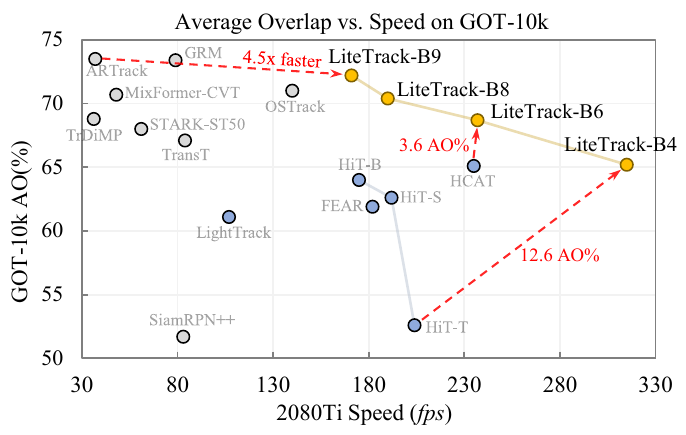}
 \setlength{\abovecaptionskip}{1pt}
 \caption
 {
	Performance comparison of LiteTrack against state-of-the-art trackers on GOT-10k in terms of Average Overlap and RTX 2080Ti Speed. \bkdot ~and \bldot ~reperesent for non-real-time and real-time trackers respectively, based on Nvidia Jetson Orin NX speed (see Tab.\ref{tab:sota}). Our LiteTrack (\yldot) family offers comparable accuracy to all other trackers, significantly outpacing them in inference speed. Notably, LiteTrack-B4 achieves over 300 \emph{fps} on 2080Ti and 100 \emph{fps} (ONNX) on edge device. Notice that our LiteTrack delivering the best real-time accuracy trained without extra data, unlike the other efficient trackers.
 }
 \label{fig:title}
 \vspace{-8mm}
\end{figure}

\begin{figure}[t] 
	\centering
	\includegraphics[width=0.98\linewidth]{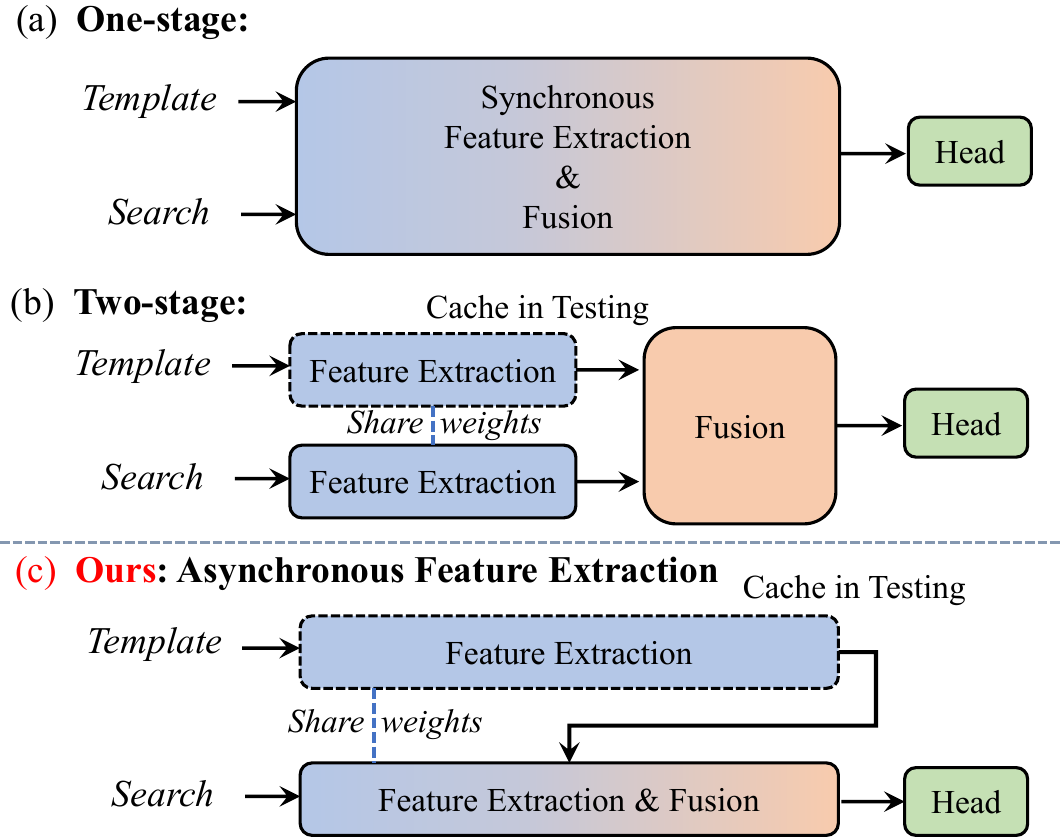}
	\setlength{\abovecaptionskip}{7pt}
	\caption
	{
	   Comparison of the popular architectures for visual tracking.  Our method (c) is able to cache the template features like two-stage (b) in testing and also enjoy the powerful pretrain technique like one-stage method (a).
	}
	\label{fig:arch_vs}
	\vspace{-2mm}
\end{figure}

\begin{figure}[t] 
	\centering
	\includegraphics[width=0.98\linewidth]{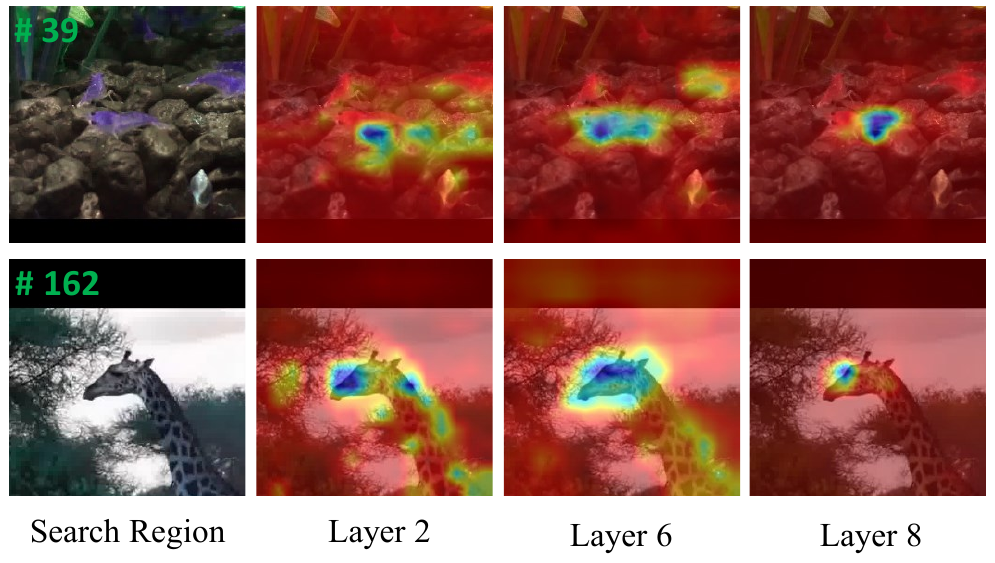}
	\setlength{\abovecaptionskip}{2pt}
	
	\caption
	{
	   Visualization of attention map (average attention value over all template features attending to the search features) of 2nd , 6th  and 8th layer in the twelve-layer enoder of JNTrack~\cite{JN}. The model focuses nearly precisely on the target even in the early stage of the encoder. 
	}
	\label{fig:early_attn}
    \vspace{-6mm}
\end{figure}

The one-stage structure has gained popularity in tracking applications~\cite{ostrack,simtrack,sbt,mixformer}. This structure combines feature extraction and fusion as a joint process as pictured in Fig.~\ref{fig:arch_vs} (a), leveraging the capabilities of the transformer network, especially the ViT~\cite{ViT} that has been pre-trained by mask-image-modeling(MIM)~\cite{MAE,CAE}. 
Conversely, two-stage trackers~\cite{Stark,TransT,SiamMask}, operating by sequentially extracting features and then fusing them, benefit from caching template features during the testing phase, as shown in Fig.~\ref{fig:arch_vs} (b).
However, the two-stage trackers who extract feature first then perform feature fusion, can cache the template feature during testing, while the one-stage trackers can not. 
Even though most of one-stage trackers are running faster than the two-stage, we can further accelerate the former by the similar caching technique.  
Inspired by ViTDet~\cite{ViTDet}, we find that only the last layer of template feature is sufficient and better for fusion with the search feature of various earlier layer, which can be cached in the testing like two-stage trackers.
Therefore, this naturally decides our overall design: the feature extraction of template is performed first and individually, then the extracted last-layer template features interact with the feature extraction of the search region, as shown in Fig.~\ref{fig:arch_vs}(c).

Traditional efficient trackers have primarily sought to achieve faster runtimes by directly incorporating an initially lightweight-designed network as their backbone. These lightweight networks are designed for efficiency, which results in relatively mediocre performance in their upstream tasks like image classification. Consequently, when such networks are utilized in visual tracking, their performance leaves much to be desired.
In contrast, our approach derive efficient model by scaling down a high-performing heavy tracker, instead of starting with a lightweight architecture. This strategy is inspired by our observation, as depicted in Fig.\ref{fig:early_attn}, that early layers pay sufficient attention to the target. By pruning network layers and integrating our novel asynchronous feature extraction technique, we ensure only a marginal drop in performance even when multiple layers are excised. Consequently, LiteTrack not only rivals the performance of its heavyweight peers but also competes in runtime with lightweight models, presenting an optimal trade-off. Fig.\ref{fig:title} reinforces this assertion, showcasing LiteTrack's commendable performance on the challenging GOT-10k\cite{GOT10K} benchmark, standing shoulder-to-shoulder with state-of-the-art (SOTA) trackers.

Our contributions are summarized as follows: 
\begin{itemize}

	\item  A efficient tracking achitechture which feature extractions of template and search region are asynchronous is proposed for reducing redundant computation.
	
	\item  A novel scaling principle of tracking model is introduced by adjusting encoder layers for trade-off between accuracy and speed. 
	
	\item  Comprehensive evaluations on authoritative generic visual tracking
	benchmarks have validated the excellent performance of LiteTrack compared with other SOTA trackers.  Edge device deployment are tested with promising performance, demonstrating the superior effectiveness of LiteTrack on robotics applicability.
\end{itemize}

\begin{figure*}[htbp]
	\centering
	\includegraphics[width=0.92\linewidth]{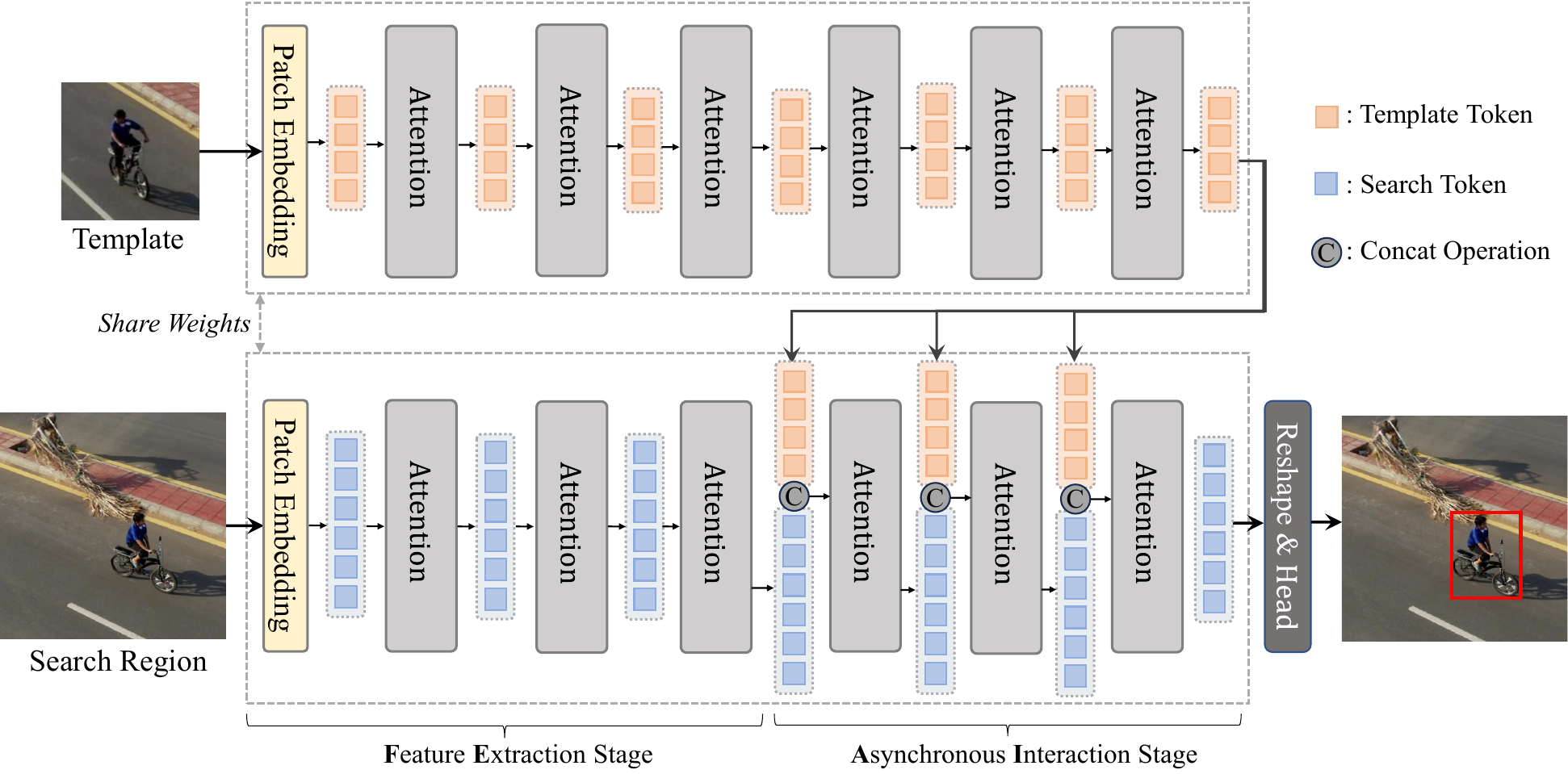}
	\caption{Overview of the proposed LiteTrack-B6 tracker, consist of 3 layers in feature extraction (FE) stage and 3 layers in asynchronous interaction (AI) stage. For simplicity, we omit the position enocding, skip connetion and MLP in the figure. Two branchs of network for template and search region share the same weights.}
	\label{fig:arch}
	\vspace{-6mm}
\end{figure*}
\vspace{-2mm}
\section{Related Works}
\vspace{-1mm}
\subsection{Visual Tracking with Transformers.}
Visual tracking has seen the rise of Siamese-based methods~\cite{SiameseFC,SINT,SiameseRPN,SiamMask,SiamRPNplusplus,SiamFC++,SiamCAR,SiamBAN,Deeper-wider-SiamRPN} that typically employ dual-backbone networks with shared parameters. They have been instrumental in the field due to their efficiency in feature extraction of the template and search region images. Further advancements introduced transformers~\cite{2017Attention} into the tracking communaty~\cite{TransT, wang2021transformer,Stark,liu2021swin,ToMP,CSWinTT,AiATrack}, leveraging them for feature interaction training from scratch. The emergence of the one-stream framework~\cite{mixformer, sbt, simtrack,ostrack,GRM} showcased improved performance by integrating feature extraction and fusion within the backbone network, enjoying the powerful mask-image-modeling (MIM) pretraining method~\cite{clip21,CAE,MAE}. Despite their effectiveness, these methods, tailored for powerful GPUs, often falter in speed on edge devices.  In response, our research incorporates the last layer feature of the template directly into the search region's feature extraction, provide a simliar cache-in-testing ability like two-stage tracker while also enjoying the powerful pretraining. 

\subsection{Lightweight Trackers.}
Efficiency in tracking is crucial for practical robotics applications, especially on edge devices. Early methods such as ECO~\cite{danelljan2017eco} and ATOM~\cite{ATOM} focused on real-time operation but didn't achieve the accuracy levels of newer trackers. Recent advancements~\cite{yan2021lighttrack,borsuk2022fear,HiT} have employed lightweight-designed backbones for efficient real-time tracking. However, these solutions still show a performance gap when compared to SOTA heavyweight trackers~\cite{ostrack, Stark, mixformer}. There have been efforts to refine these advanced trackers: OSTrack~\cite{ostrack} considered pruning non-essential features unrelated to the foreground, while SimTrack~\cite{simtrack} suggested removing the last four layers to cut computational costs. Despite these modifications, there remains a lack of deep exploration into true real-time lightweight tracking architectures. Our proposed LiteTrack fills this gap, combining efficiency and performance for effective tracking on edge devices.

\section{Proposed Method}
This section presents the LiteTrack method in detail.
First, we briefly overview our LiteTrack framework. Then, we depict the asynchronous featrue extraction process, layer pruning of our model, and the head network plus with the training obejctive. 

\subsection{Overview}
As shown in Fig.~\ref{fig:arch}, LiteTrack is a combination of one-stage and two-stage tracking framework consisting of two components: the lightweight transformer encoder and the head network. 
The template of target to be tracked is fed into the lightweight transformer encoder first for feature extraction individually.   Then the image of search region is also fed into the same encoder but only the first n layers.  We call these first n layers as \textit{Featrure Extraction Stage} (FE).  Next, the extracted template features of last layer together with the intermediate search features after the featrure extraction stage, are fed as a concatenated sequence into the remaining encoder layers.  We call these final layers as \textit{Asynchronous Interaction Stage} (AI).  Finally, only the part belongs to the search of the final sequence are selected and flatten to 2D feature map, and fed into the head network for the tracking result.

\subsection{Asynchronous Featrure Extraction}
The so-called \textit{asynchronous} means that we extract the template feature first and then the search feature.  The feature extraction is done by the transformer encoder, in which each layer mainly constists of multi-head attention.  Specifically, for a template image $\mathbf{Z} \in {\mathbb{R}}^{3 \times {H_{z}} \times {W_{z}}}$, the pacth embedding layer transform the image into sequence of tokens  $\mathbf{Z_{p}} \in {\mathbb{R}}^{C \times {\frac{H_{z}}{16}} \times {\frac{W_{z}}{16}}}$. In each layer of the transofmer encoder, the main operation is multi-head self attention: 
\begin{equation}
	\label{eq-m-h-a-z}
	\begin{aligned}
		 {\rm{Attn}_z}
		 =& {\rm{softmax}}(\frac{\mathbf{Q}_z\mathbf{K}_z^\top}{\sqrt{d_k}})\mathbf{V}_z. 
	\end{aligned}
\end{equation}
The featrue extraction of template is performed by a serially stacked encoder layers.
For a search image $\mathbf{X} \in {\mathbb{R}}^{3 \times {H_{x}} \times {W_{x}}}$, the same pacth embedding layer also transform the image into sequence of tokens $\mathbf{X_{p}} \in {\mathbb{R}}^{C \times {\frac{H_{x}}{16}} \times {\frac{W_{x}}{16}}}$.
In the \textit{feature extraction stage}, the search tokens only attent to itself in the multi-head attention operation:
\begin{equation}
	\label{eq-m-h-a-x}
	\begin{aligned}
		 {\rm{Attn}_x}
		 =& {\rm{softmax}}(\frac{\mathbf{Q}_x\mathbf{K}_x^\top}{\sqrt{d_k}})\mathbf{V}_x. 
	\end{aligned}
\end{equation}
In the \textit{asynchronous interaction stage}, the tokens of extracted template features of last layer concatenated together with the intermediate search features after the \textit{feature extraction stage} are concatenated, as the input of the encoder layer.  Inspired by MixFormer~\cite{mixformer}, the attention in the layers within the interaction stage is a little different from the standard self-attention: we generate queris $\mathbf{Q}$ only by the search features. Thus the attention process can be written as:
\begin{equation}
	\label{eq-m-h-a-xz}
	\begin{aligned}
		 {\rm{Attn}_{xz}}
		 =& {\rm{softmax}}(\frac{\mathbf{Q}\mathbf{K}^\top}{\sqrt{d_k}})\mathbf{V}\\
		 =& {\rm{softmax}}(\frac{[\mathbf{Q}_x][\mathbf{K}_x^\top;\mathbf{K}_z^\top]}{\sqrt{d_k}})\mathbf[{V}_x;{V}_z]. 
	\end{aligned}
\end{equation}
Though the attention computation are different between two stages, the parameters of the networks are still in the same structure. Therefore the template branch and search branch of the network can share the weight.    

\noindent \textbf{\textit{Analysis:}} Our method works depending on one critical development of recent trackers: the application of homogeneous structure backbone such as the ViT~\cite{ViT} encoder.
The chanel of the features, or the dimension of the tokens in context of the transformer, remains unchange during the layers of encoder, therefore the template features can interact with the search features from any intermediate encoder layer.
\begin{figure}[t] 
	\centering
	\includegraphics[width=0.98\linewidth]{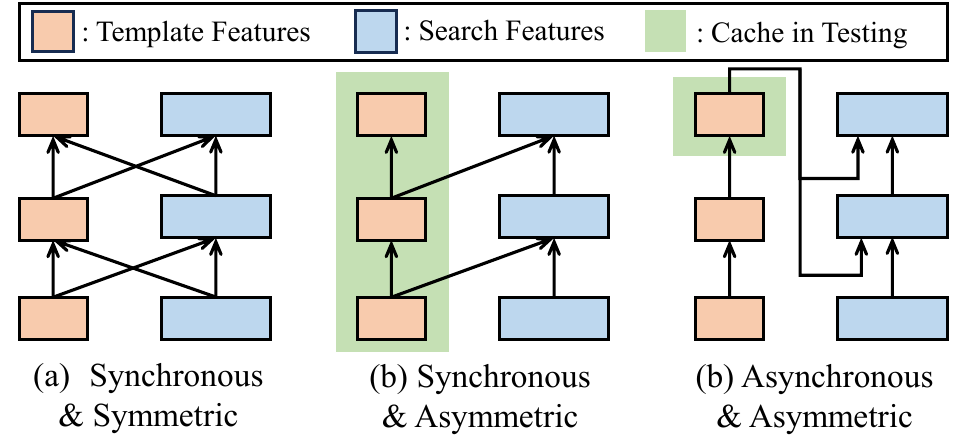}
	\setlength{\abovecaptionskip}{7pt}
	\caption
	{
		Comparative architectures of recent one-stage one-stream trackers versus our proposed method. (a) Symmetric interaction: both template and search region features engage layer by layer. (b) Template-only updates: features of the template update without search region interactions. (c) Last-layer interaction: only the final-layer template feature engages with the search region.
	}
	\label{fig:extraction_vs}
    \vspace{-2mm}
\end{figure}
This introduce our first model scaling principle: adjusting the layers for the feature extraction stage and the asynchronous stage for the accuracy-speed trade-off as discussed in Sec.~\ref{sec:configuration}.
During testing, synchronous and symmetric feature extraction methods, such as OSTrack~\cite{ostrack} shown in Fig.\ref{fig:extraction_vs}(a), often result in redundant computations for the template. Given that the template, typically the initial frame of a video sequence, remains unchanged, its feature also remains constant. By caching these template features, we eliminate unnecessary computations during testing. In contrast to MixFormer, which caches every layer of template features as depicted in Fig.\ref{fig:extraction_vs}(b), our method conserves memory by only storing the last layer of template features, as shown in Fig.~\ref{fig:extraction_vs}(c).

\subsection{Layer Pruning}
\begin{figure}[t] 
	\centering
	\includegraphics[width=0.99\linewidth, trim=0 17 0 0,clip]{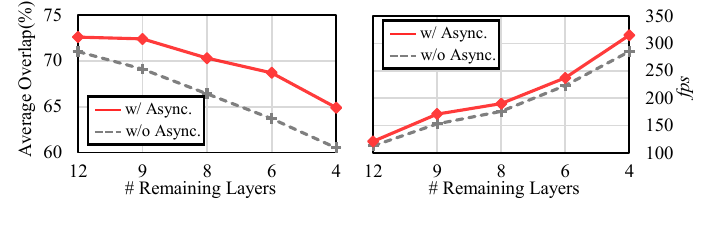}
	\caption
	{
	   Performance (left) and speed (right) vs. number of pruning layers on GOT-10k benchmark.  The baseline without asynchronous feature extraction is bulit upon OSTrack~\cite{ostrack}. 
	}
	\label{fig:pruning}
    \vspace{-4mm}
\end{figure}
In the pursuit of enhancing object tracking performance, deep neural networks have grown increasingly complex, often at the expense of computational efficiency. Layer pruning offers an avenue to mitigate this by systematically reducing the number of layers in the network. Starting with a 12-layer ViT encoder, we adopted a top-down pruning strategy, progressively eliminating layers and assessing performance against a baseline.  As illustrated in Fig.\ref{fig:pruning}, the performance dropped with the layers pruned while the speed rise up significantly.  However, when paired with asynchronous feature extraction,  the decreasing of performance by the layer pruning is moderated.  For example, our 9-layer variant combined with asynchronous as outperformed its 12-layer counterpart as shown in Fig.\ref{fig:pruning}. 

\subsection{Head and Training Objective}
\label{sec:training_and_inference}
We employ the center head~\cite{ostrack} for prediction, which consists of three convolutional branches for center classification, offset regression and size regression, respectively. The center classification
branch outputs a centerness score map, where each score represents the confidence of the target center locating at the corresponding position. The prediction of offset regression branch is for the discretization error of the center. The size regression branch predicts the height and width of the target. The position with the highest confidence in the center score map is selected as the target position and the corresponding regressed coordinates are used to compute a bounding box as the final prediction.

We apply the weighted focal loss~\cite{focal} for classification. For localization, we combine $\ell_1$ loss and the generalized GIoU loss~\cite{GIoU} as the training objective. The overall loss function can be formulated as
\begin{equation}
	\label{equ-loss-loc}
	\begin{aligned}
		\mathcal{L}=\mathcal{L}_{\operatorname{focal}} + \lambda_{G}\mathcal{L}_{\operatorname{GIoU}}+\lambda_{l}\mathcal{L}_l,
	\end{aligned}
\end{equation}
where $\lambda_{G}=2$ and $\lambda_{l}=5$ are trade-off weights following ~\cite{ostrack} to balance optimization.

\section{Experiments}
\subsection{Implementation Details}
\begin{table}[t]\footnotesize
    \centering
    \resizebox{\linewidth}{!}{
        
        \begin{tabular}{c | l | c c c c}
            \toprule
            \multicolumn{2}{c|}{Model}     & LiteT-B9        & LiteT-B8 & LiteT-B6 & LiteT-B4        \\
            \midrule[0.5pt]
            \multicolumn{2}{c|}{\# FE Layers} & 6               & 6        & 3        & 2               \\
            \multicolumn{2}{c|}{\# AI Layers} & 3               & 2        & 3        & 2               \\
            \midrule[0.5pt]
            \multirow{3}*{\emph{fps}}                      & 2080Ti (PyTorch)         & 171      & 190      & 237      & 315  \\
                               & OrinNX  (PyTorch)          & 21       & 25       & 31       & 44   \\
                               & OrinNX  (ONNX fp16)          & 64       & 70       & 82       & 102   \\
            \midrule[0.5pt]
            \multicolumn{2}{c|}{MACs(G)}   & 14.17           & 12.77    & 10.09    & 6.78            \\
            \midrule[0.5pt]
            \multicolumn{2}{c|}{Params(M)} & 54.92           & 49.60    & 38.97    & 26.18           \\
            \bottomrule
        \end{tabular}
    }
    \caption{Details and variants of our LiteTrack model. FE and AI denote for Feature Extraction and Asynchronous Interaction, respectively.
    }
    \label{tab:speed}
    \vspace{-5mm}
\end{table}
\begin{table*}[t]\footnotesize
    \centering
    \setlength{\tabcolsep}{2mm}{
        \resizebox{\textwidth}{!}{
            \begin{tabular}{c|l|c|c ccc c ccc c ccc c cc}
                \toprule
                                                              & \multirow{2}*{Method}               & \multirow{2}*{Source} & \multicolumn{3}{c}{TrackingNet~\cite{trackingnet}} &                         & \multicolumn{3}{c}{LaSOT~\cite{LaSOT}} &     & \multicolumn{3}{c}{GOT-10k$^*$~\cite{GOT10K}} &                         & \multicolumn{2}{c}{Speed (\emph{fps})}                                                                                                              \\
                \cline{4-6}
                \cline{8-10}
                \cline{12-14}
                \cline{16-17}
                                                              &                                     &                       & AUC                                                & P$_{Norm}$              & P                                      &     & AUC                                           & P$_{Norm}$              & P                                      &     & AO                      & SR$_{0.5}$              & SR$_{0.75}$             &     & 2080Ti  & OrinNX \\
                \midrule[0.5pt]
                \multirow{9}*{\rotatebox{90}{Real-time}}      & \my LiteTrack-B9                    & \my Ours              & \my\sotaBest{82.4}                                 & \my\sotaBest{87.3}      & \my\sotaBest{80.4}                     & \my & \my\sotaBest{67.0}                            & \my\sotaBest{77.0}      & \my\sotaBest{72.7}                     & \my & \my\sotaBest{72.2}      & \my\sotaBest{82.3}      & \my\sotaBest{69.3}      & \my & \my 171 & \my 21 \\
                                                              & \my LiteTrack-B8                    & \my Ours              & \my\sotaSecond{81.4}                               & \my\sotaSecond{86.4}    & \my\sotaSecond{79.4}                   & \my & \my\sotaSecond{66.4}                          & \my\sotaSecond{76.4}    & \my\sotaSecond{71.4}                   & \my & \my\sotaSecond{70.4}    & \my\sotaSecond{80.1}    & \my\sotaSecond{66.4}    & \my & \my 190 & \my 25 \\
                                                              & \my LiteTrack-B6                    & \my Ours              & \my\sotaThird{80.8}                                & \my\sotaThird{85.7}     & \my\sotaThird{78.2}                    & \my & \my\sotaThird{64.6}                           & \my\sotaThird{73.9}     & \my\sotaThird{68.9}                    & \my & \my\sotaThird{68.7}     & \my\sotaThird{78.2}     & \my\sotaThird{64.2}     & \my & \my 237 & \my 31 \\
                                                              & \my LiteTrack-B4                    & \my Ours              & \my 79.9                                           & \my 84.9                & \my 76.6                               & \my & \my 62.5                                      & \my 72.1                & \my 65.7                               & \my & \my 65.2                & \my 74.7                & \my 57.7                & \my & \my 315 & \my 44 \\
                                                              & HiT-Base\footnotemark[1]~\cite{HiT} & ICCV'23               & 80.0                                               & 84.4                    & 77.3                                   &     & \sotaThird{64.6}                              & 73.3                    & 68.1                                   &     & \nGOT{64.0}             & \nGOT{72.1}             & \nGOT{58.1}             &     & 175     & -      \\
                                                              & E.T.Track~\cite{ETTrack}            & WACV'23               & 75.0                                               & 80.3                    & 70.6                                   &     & 59.1                                          & -                       & -                                      &     & -                       & -                       & -                       &     & 67      & 21     \\
                                                              & FEAR-XS~\cite{borsuk2022fear}       & ECCV'22               & -                                                  & -                       & -                                      &     & 53.5                                          & -                       & 54.5                                   &     & \nGOT{61.9}             & \nGOT{72.2}             & -                       &     & 182     & 50     \\
                                                              & HCAT~\cite{chen2022efficient}       & ECCVW'22              & 76.6                                               & 82.6                    & 72.9                                   &     & 59.3                                          & 68.7                    & 61.0                                   &     & \nGOT{65.1}             & \nGOT{76.5}             & \nGOT{56.7}             &     & 235     & 34     \\
                                                              & LightTrack~\cite{yan2021lighttrack} & CVPR'21               & 72.5                                               & 77.8                    & 69.5                                   &     & 53.8                                          & -                       & 53.7                                   &     & \nGOT{61.1}             & \nGOT{71.0}             & -                       &     & 107     & 38     \\
                                                              & HiFT~\cite{hift,2023SOTsurvey}      & ICCV'21               & 66.7                                               & 73.8                    & 60.9                                   &     & 45.1                                          & 52.7                    & 42.1                                   &     & -                       & -                       & -                       &     & 230     & 50     \\
                                                              & ECO~\cite{danelljan2017eco}         & CVPR'17               & 55.4                                               & 61.8                    & 49.2                                   &     & 32.4                                          & 33.8                    & 30.1                                   &     & 39.5                    & 40.7                    & 17.0                    &     & 113     & 22   \\
                \midrule[0.5pt]
                \multirow{10}*{\rotatebox{90}{Non-real-time}} & ARTrack-256~\cite{ARTrack}          & CVPR'23               & \underline{\emph{84.2}}                            & \emph{\underline{88.7}} & 83.5                                   &     & \emph{\underline{70.4}}                       & \emph{\underline{79.5}} & \emph{\underline{76.6}}                &     & \emph{\underline{73.5}} & 82.2                    & \emph{\underline{70.9}} &     & 37      & 6      \\
                                                              & GRM-256~\cite{GRM}                  & CVPR'23               & 84.0                                               & \emph{\underline{88.7}} & 83.3                                   &     & 69.9                                          & 79.3                    & 75.8                                   &     & 73.4                    & \emph{\underline{82.9}} & 70.4                    &     & 79      & 10     \\
                                                              & OSTrack-256~\cite{ostrack}          & ECCV'22               & 83.1                                               & 87.8                    & 82.0                                   &     & 69.1                                          & 78.7                    & 75.2                                   &     & 71.0                    & 80.4                    & 68.2                    &     & 140     & 18     \\
                                                              & MixFormer~\cite{mixformer}          & CVPR'22               & 83.1                                               & 88.1                    & 81.6                                   &     & 69.2                                          & 78.7                    & 74.7                                   &     & 70.7                    & 80.0                    & 67.8                    &     & 48      & 12     \\
                                                              & Sim-B/16~\cite{simtrack}            & ECCV'22               & 82.3                                               & 86.5                    & -                                      &     & 69.3                                          & 78.5                    & -                                      &     & 68.6                    & 78.9                    & 62.4                    &     & 131     & 15     \\
                                                              & STARK-ST50~\cite{Stark}             & ICCV'21               & 81.3                                               & 86.1                    & -                                      &     & 66.6                                          & -                       & -                                      &     & 68.0                    & 77.7                    & 62.3                    &     & 61      & 13     \\
                                                              & TransT~\cite{TransT}                & CVPR'21               & 81.4                                               & 86.7                    & 80.3                                   &     & 64.9                                          & 73.8                    & 69.0                                   &     & 67.1                    & 76.8                    & 60.9                    &     & 84      & 13     \\
                                                              & TrDiMP~\cite{wang2021transformer}   & CVPR'21               & 78.4                                               & 83.3                    & 73.1                                   &     & 63.9                                          & -                       & 61.4                                   &     & 67.1                    & 77.7                    & 58.3                    &     & 36      & 6      \\
                                                              & PrDiMP~\cite{PrDiMP}                & CVPR'20               & 75.8                                               & 81.6                    & 70.4                                   &     & 59.8                                          & 68.8                    & 60.8                                   &     & 63.4                    & 73.8                    & 54.3                    &     & 47      & 12     \\
                                                              & DiMP~\cite{DiMP}                    & ICCV'19               & 74.0                                               & 80.1                    & 68.7                                   &     & 56.9                                          & 65.0                    & 56.7                                   &     & 61.1                    & 71.7                    & 49.2                    &     & 100     & 16     \\
                                                              & SiamRPN++~\cite{SiamRPNplusplus}    & CVPR'19               & 73.3                                               & 80.0                    & 69.4                                   &     & 49.6                                          & 56.9                    & 49.1                                   &     & 51.7                    & 61.6                    & 32.5                    &     & 83      & 16     \\
                                                              & ATOM~\cite{ATOM}                    & CVPR'19               & 70.3                                               & 77.1                    & 64.8                                   &     & 51.5                                          & 57.6                    & 50.5                                   &     & \nGOT{55.6}             & \nGOT{63.4}             & \nGOT{40.2}             &     & 175     & 15     \\
                \bottomrule
            \end{tabular}
        }}
    \caption{State-of-the-art comparison on TrackingNet~\cite{trackingnet}, LaSOT~\cite{LaSOT}, and GOT-10k~\cite{GOT10K} benchmarks.  The best three real-time results are shown in \sotaBest{red}, \sotaSecond{blue} and \sotaThird{green} fonts, and the best non-real-time results are shown in \underline{\emph{underline}} font. * denotes results on GOT-10k obtained following the official one-shot protocol, with \textcolor{gray!60}{gray} font indicating training using extra data.
    }
    \label{tab:sota}
    \vspace{-4mm}
\end{table*}
\begin{table}[t]\small
    \centering
    \setlength{\tabcolsep}{4mm}{
        \resizebox{1\linewidth}{!}{
            \begin{tabular}{l|ccc}
                \toprule
                \multirow{2}*{Method}               & NFS~\cite{NFS}        & UAV123~\cite{UAV}     & VOT'21~\cite{vot2021}  \\
                                                    & AUC                   & AUC                   & EAO                    \\
                \midrule[0.5pt]

                \my LiteTrack-B9                    & \my \sotaBest{65.4}   & \my \sotaBest{67.7}   & \my \sotaBest{0.269}   \\
                \my LiteTrack-B8                    & \my \sotaSecond{64.6} & \my \sotaSecond{67.1} & \my \sotaSecond{0.261} \\
                \my LiteTrack-B6                    & \my \sotaThird{64.4}  & \my {66.2}            & \my \sotaThird{0.254}  \\
                \my LiteTrack-B4                    & \my 63.4              & \my \sotaThird{66.4}  & \my 0.251              \\
                HiT-Base~\cite{HiT}                 & 63.6                  & 65.6                  & 0.252                  \\
                HCAT~\cite{chen2022efficient}       & 63.5                  & 62.7                  & -                      \\
                FEAR~\cite{borsuk2022fear}          & 61.4                  & -                     & 0.250                  \\
                E.T.Track~\cite{ETTrack}            & 59.0                  & 62.3                  & 0.224                  \\
                LightTrack~\cite{yan2021lighttrack} & 55.3                  & 62.5                  & 0.225                  \\
                HiFT~\cite{hift}                    & -                     & 58.9                  & -                      \\
                ECO~\cite{danelljan2017eco}         & 46.6                  & 53.2                  & -                      \\
                \bottomrule
            \end{tabular}
        }}
    \caption{Comparison with state-of-the-art real-time trackers on additional benchmarks.  We report the results on the NFS~\cite{NFS} and UAV123~\cite{UAV} in AUC score, and the EAO for VOT2021~\cite{vot2021} real-time benchmark.}
    \label{tab:sota-small}
    \vspace{-7mm}
\end{table}
\textit{Model Variants.} We use ViT-B~\cite{ViT} pretrained by CAE~\cite{CAE} as our backbone. We develop four variants of LiteTrack with using different layers of the transformer, as elaborated in Tab.~\ref{tab:speed}. We adopt first 9, 8, 6 and 4 layers of the 12-layer ViT-B model for LiteTrack-B9, LiteTrack-B8, LiteTrack-B6, and LiteTrack-B4, respectively. FE layers and AI layers in the table denote for the layers used in feature extraction stage and asynchronous interaction stage, respectively.
In addition, Tab.~\ref{tab:speed} reports model parameters, Multiply-Accumulate operations (MACs), and inference speed on multiple environments.  The running setups are detailed decribed in Sec~\ref{sec:setup}.

\textit{Training. }   
For the GOT-10k~\cite{GOT10K} benchmark, we only use the training split of GOT-10k following the one-shot protocols and train the model for 100 epochs.  
For the other benchmarks, the training splits of GOT-10k, COCO~\cite{COCO}, LaSOT~\cite{LaSOT} and TrackingNet~\cite{trackingnet} are used for training in 300 epochs.
For video datasets, we sample the image pair from a random video sequence. For the image dataset COCO, we randomly select an image and apply data augmentations to generate an image pair. 
Common data augmentations such as scaling, translation, and jittering are applied on the image pair. 
The search region and the template are obtained by expanding the target box by a factor of 4 and 2, respectively.
The optimizer is the AdamW optimizer~\cite{AdamW}, with the weight decay of 1e-4. 
The initial learning rate of the encoder and the decoder are 4e-5 and 4e-4, respectively.
We reduce the learning rate to 10\% in the last 20\% epochs.  
Each GPU holds 64 image-pairs.
Training of LiteTrack-B8 takes about 9 hours for GOT-10k and 24 hours for the other benchmarks on one RTX 3090 GPU, and the lighter model trains faster.

\textit{Testing.} 
During inference, the template is initialized in the first frame of a video sequence. 
For each subsequent frame, the search region is cropped based on the target's bounding box of the previous frame.
We adopt Hanning window penalty to utilize positional prior like scale change and motion smoothness in tracking, following the common practice~\cite{CSWinTT, ostrack}. The output scores are simply element-wise multiplied by the Hanning window with the same size, and we choose the box with the highest multiplied score as the target box.

\footnotetext[1]{The \emph{fps} on GPU of HiT~\cite{HiT} is directly from their paper due to limitated access to their code. We consider HiT as a real-time tracker based on their reported results on the AGX Xavier, which has a performance near to our Orin NX. For a hardware comparison, see the benchmarks on \href{https://developer.nvidia.com/embedded/jetson-benchmarks}{https://developer.nvidia.com/embedded/jetson-benchmarks}.}

\vspace{-2mm}
\subsection{State-of-the-art Comparisons}
\vspace{-1mm}
\label{sec:setup}
LiteTrack is benchmarked against state-of-the-art trackers, both real-time and non-real-time, across six tracking datasets. We evaluated the speed of these trackers on two distinct platforms: an Nvidia GeForce RTX 2080Ti GPU (with an Intel i5-11400F CPU) and an Nvidia Jetson Orin NX 16GB edge device. For these tests, we utilized PyTorch 1.12.0 on the former and PyTorch 2.0.0 @ JetPack 5.1 on the latter.  
Trackers are categorized into real-time and non-real-time based on their PyTorch speed on the Orin NX device, following the 20 \emph{fps} real-time setting of VOT~\cite{vot2020}.  Detailed comparative results are showcased in Tables~\ref{tab:sota} and~\ref{tab:sota-small}.  We also report our tracker's speed accelerated with ONNX fp16 in Tab.~\ref{tab:speed}.

\textit{GOT-10k.} GOT-10k~\cite{GOT10K} is a large-scale and challenging dataset that contains 10k training sequences and 180 test sequences which are zero-overlapping in classes.  The official one-shot protocol reuqires the evaluated tracker training without extra data, which encourages the model designed for challenging scenes like unseen objects. We report the Average Overlap (AO), Success Rate over a overlapping rate of 50\% (SR$_{0.5}$) and the same one over 75\%   (SR$_{0.75}$) obtained by submitting the result to the official evaluation server.  As shown in Table~\ref{tab:sota}, LiteTrack-B9 achieves the best real-time results of 72.2\% AO score, which is also competitive to the best non-real-time tracker ARTrack-256~\cite{ARTrack} (73.5\% AO score). Our LiteTrack-B4 surpassing all the real-time trackers with AO score of 65.2\%, even though our trackers are trained without extra data. 

\textit{TrackingNet.} TrackingNet~\cite{trackingnet} is a large-scale dataset containing a variety of situations in natural scenes and multiple categories, and its test set includes 511 video sequences. We report the Area Under Curve (AUC), Normalized Precision (P$_{Norm}$) and Precision (P) obtained by submitting the tracking result to the official evaluation server.  As reported in Table~\ref{tab:sota}, LiteTrack series achieve competitive results compared with the previous real-time trackers. LiteTrack-B9 gets the best AUC of 82.4\%, surpassing the previous best real-time tracker HiT-Base~\cite{HiT} by 2.4\%. Compared to non-real-time tracker ARTrack~\cite{ARTrack}, LiteTrack-B9 achieves comparable performance to it in AUC (82.4 $vs$. 84.2) while being $4.5 \times$ faster on the GPU and $6 \times$ faster on the Jetson edge platform.

\textit{LaSOT.} LaSOT~\cite{LaSOT} is a large-scale, long-term dataset containing 1400 video sequences, with 1120 training videos and 280 test videos.  We report the same matrices as in TrackingNet evaluated by PyTracking\footnote[2]{\href{https://github.com/visionml/pytracking}{https://github.com/visionml/pytracking}} tools. The results on LaSOT are shown in Table~\ref{tab:sota}. LiteTrack-B9 achieves the best real-time results of 67.0\%, 77.0\%, and 72.7\% in AUC, P$_{Norm}$, and P, respectively. LiteTrack-B8 and LiteTrack-B6 achieves the second-best and the third-best AUC score. 
Compared with the recent efficient tracker HiT-Base~\cite{HiT}, LiteTrack-B9 outperform it by 2.4\% in AUC. 

\textit{NFS, UAV123 and VOT2021.} On the NFS dataset~\cite{NFS}, known for its fast-moving objects spanning 100 video sequences, our LiteTrack variants B9, B8, and B6 emerge as the top three in real-time performance as highlighted in Table~\ref{tab:sota-small}. Meanwhile, on the UAV123 dataset~\cite{UAV}, which features 123 video clips from low-altitude UAVs, even our fastest LiteTrack-B4 takes the lead among real-time trackers with an AUC score of 66.4\%, surpassing competitors such as HiT~\cite{HiT} and HCAT~\cite{chen2022efficient} by margins of 0.8\% and 3.7\%, respectively. Similarly, our VOT-2021 real-time experiments on the VOT2021 benchmark~\cite{vot2021} witnessed LiteTrack-B9 achieving the highest EAO score of 26.9\% among real-time trackers, as tabulated in Table~\ref{tab:sota-small}.

\subsection{Ablation Study and Visualization}
\begin{table}
    \centering
    \resizebox{\linewidth}{!}{
        \begin{tabular}{c|l|cc}
            \toprule
            \# & \multicolumn{1}{c|}{Method}         & AO   & {\emph{fps}} \\
            \midrule[0.5pt]
            1  & OSTrack(w/o CE)                     & 71.0 & 121          \\

            2  & Strong baseline(1 + CAE~\cite{CAE} pretrained)  & 71.7 & 120          \\

            3  & 2 + Last 3 layers pruned           & 69.1 & 162          \\

            4  & \myAbl 3 + Asynchronous Feature Extraction & \myAbl \textbf{72.2} &\myAbl \textbf{171}          \\
            \bottomrule
        \end{tabular}}
        \setlength{\abovecaptionskip}{4pt}
        \caption{Comparison of key components for object tracking performance. We use \colorbox{gray!20}{gray} color to denote for our final configuration.}
        \label{tab:component}
        \vspace{-3mm}
\end{table}

\begin{table}
    \centering
    \resizebox{\linewidth}{!}{
        \begin{tabular}{c|cc|ccc}
            \toprule
            \# Total                       & \# FE                         & \# AI                          & \multirow{2}{*}{AO} & \multirow{2}{*}{SR$_{0.5}$} & \multirow{2}{*}{\textit{fps}} \\
            \multicolumn{1}{l|}{~ ~Layers} & \multicolumn{1}{l}{~ ~Layers} & \multicolumn{1}{l|}{~ ~Layers} &                     &                             &                               \\
            \midrule
            \multirow{3}{*}{8}             & 6                             & 2                              & 70.3                & \textbf{80.4}               & 190                           \\
                                           & \myAbl 5                             &\myAbl 3                              & \myAbl\textbf{70.4}       & \myAbl80.1                        & \myAbl185                           \\
                                           & 0                             & 8                              & 68.3                & 77.9                        & 173                           \\
            \midrule
            \multirow{2}{*}{6}             & 4                             & 2                              & 68.0                & 77.5                        & 241                           \\
                                           & \myAbl 3                             & \myAbl3                              & \myAbl\textbf{68.7}       & \myAbl\textbf{78.2}               & \myAbl237                           \\
            \midrule
            \multirow{2}{*}{4}             & 3                             & 1                              & 64.6                & 73.7                        & 318                           \\
                                           & \myAbl 2                             & \myAbl 2                              & \myAbl\textbf{65.2}       & \myAbl\textbf{75.5}               &\myAbl 315                           \\
            \bottomrule
        \end{tabular}}
    \caption{Performance comparison based on varying ratios of feature extraction layers to asynchronous interaction layers. We use \colorbox{gray!20}{gray} color to denote our final configuration.}
    \label{tab:layer}
    \vspace{-6mm}
\end{table}

\begin{figure}[t] 
	\centering
	\includegraphics[width=0.98\linewidth]{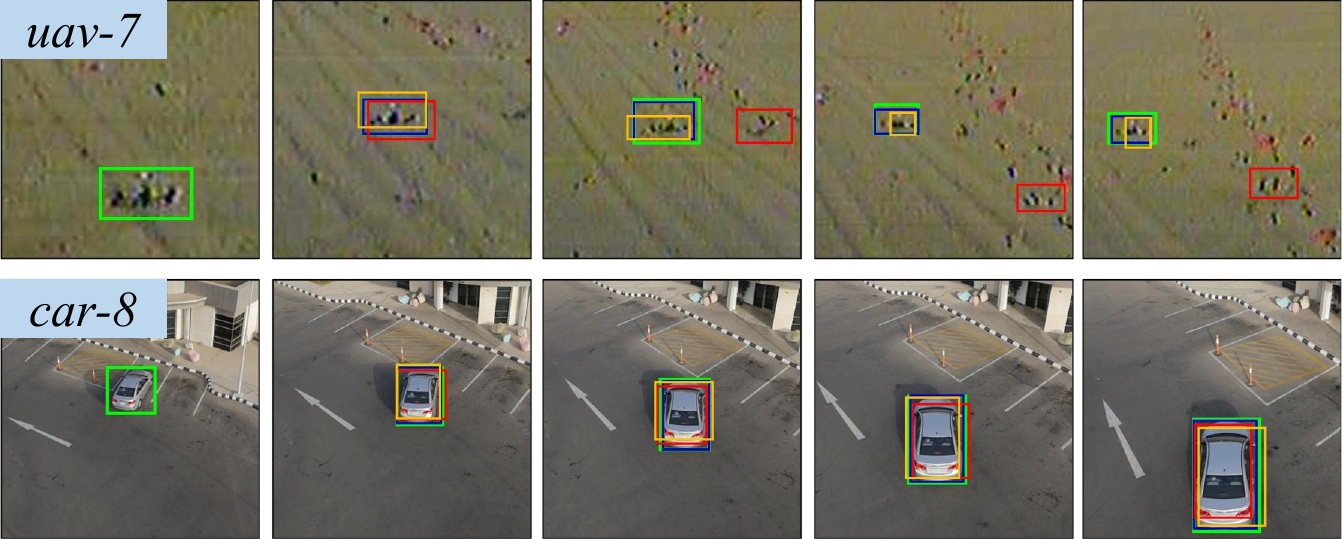}
	\setlength{\abovecaptionskip}{7pt}
	\caption
	{
		Prediction comparison from UAV123~\cite{UAV}. We use \textcolor{green}{green} lines to demonstrate the Groud Truth bounding box of the target. \textcolor{blue}{Blue} boxes  represent our LiteTrack's predictions, while \textcolor{yellow}{yellow} and \textcolor{red}{red} boxes denote the predictions of trackers HCAT~\cite{chen2022efficient} and E.T.Track~\cite{ETTrack} respectively.
	}
	\label{fig:uav}
	\vspace{-8mm}
\end{figure}


\textit{Component-wise Analysis.} The significance of our proposed methods is underscored through a comparative study built upon OSTrack~\cite{ostrack}. For setting a solid baseline, we enhanced OSTrack by substituting its MAE~\cite{MAE} pretrained weights with those of CAE~\cite{CAE}, the outcome of which is enumerated in Tab.~\ref{tab:component}, Row 2. Direct layer pruning, as seen in Row 3, led to a marked decline in performance. However, when integrated with our novel asynchronous feature extraction (Row 4), not only was the deficit recovered, but the model also achieved superior accuracy and efficiency, surpassing even the strong baseline.

\textit{Layer Configuration Analysis.} 
\label{sec:configuration}
We explored various configurations concerning the ratio of feature extraction (FE) layers to asynchronous interaction (AI) layers, as depicted in Tab.~\ref{tab:layer}. For configurations with 8 total layers, peak performance was achieved with a majority of the layers dedicated to FE. The 6-layer configuration showed comparable results, especially with an even FE-to-AI ratio. Notably, in the 4-layer configurations, a balanced 2:2 FE-to-AI setup still produced respectable results. The data highlights the model's adaptability across different layer configurations and offers insights into achieving an optimal balance between FE and AI layers.

\textit{Qualitative Results.} To better present the superiority of LiteTrack, we highlight representative scenes in Fig.~\ref{fig:uav}. In a challenging UAV tracking scenario under a noisy and jittery UAV camera feed, LiteTrack  consistently maintains its track, outperforming other trackers. Similarly, when tracking a moving car from a UAV's perspective, LiteTrack demonstrates pinpoint precision, ensuring more accurate alignment with the ground truth than competing methods. These real-world tests underscore LiteTrack's proficiency in handling diverse tracking challenges.

\vspace{-3mm}
\section{Conclusions}
\vspace{-2mm}
In this work, we've presented LiteTrack, a pioneering approach to object tracking tailored for robotics applications and edge devices. By combining layer pruning with asynchronous feature extraction, we've achieved significant improvements in both accuracy and execution speed across diverse datasets. Our results underscore LiteTrack's potential, as it not only outperforms leading real-time trackers but also addresses the constraints of computational resources often found in robotics and edge deployments. With its efficient design, LiteTrack promises to be a valuable basline for real-time robotics applications.








\bibliographystyle{IEEEtran}

\bibliography{HiT}


\end{document}